\documentclass{article}
\usepackage{spconf,amsmath,graphicx, booktabs, multirow, amsfonts, hyperref, float}
\usepackage[table,xcdraw]{xcolor}

\title{Feature-Driven Super-Resolution for Object Detection}
\name{Bin Wang$^{}$, Tao Lu$^{*}$, Yanduo Zhang$^{}$ \thanks{* is corresponding author. This work is supported by the National Natural Science Foundation of China (61502354, 61501413, 61671332, 41501505), the Natural Science Foundation of Hubei Province of China (2015CFB451, 2014CFA130, 2012FFA099, 2012FFA134, 2013CF125), Scientific Research Foundation of Wuhan Institute of Technology (K201713).}}
\address{Hubei Key Laboratory of Intelligent Robot, School of Computer Science and Engineering \\
Wuhan Institute of Technology, Wuhan, China, 430073.}

\begin{document}
	%
	\maketitle
	\begin{abstract}
Although some convolutional neural networks (CNNs) based super-resolution (SR) algorithms yield good visual performances on single images recently. Most of them focus on perfect perceptual quality but ignore specific needs of subsequent detection task.
This paper proposes a simple but powerful feature-driven super-resolution (FDSR) to improve the detection performance of low-resolution (LR) images.
First, the proposed method uses feature-domain prior which extracts from an existing detector backbone to guide the HR image reconstruction.
Then, with the aligned features, FDSR update SR parameters for better detection performance. Comparing with some state-of-the-art SR algorithms with 4$\times$ scale factor, FDSR outperforms the detection performance mAP on MS COCO validation, VOC2007 databases with good generalization to other detection networks.
	\end{abstract}
	\begin{keywords}
Feature-Driven, Super-Resolution, Object Detection
	\end{keywords}

	\vspace{-0.3cm}
	\section{Introduction}
	\vspace{-0.2cm}
	\label{sec:intro}
	
	\begin{figure}[t]
		
		\begin{minipage}[b]{0.24\linewidth}
			\centering
			\centerline{0.50/0.60}
			\centerline{\includegraphics[width=2.15cm]{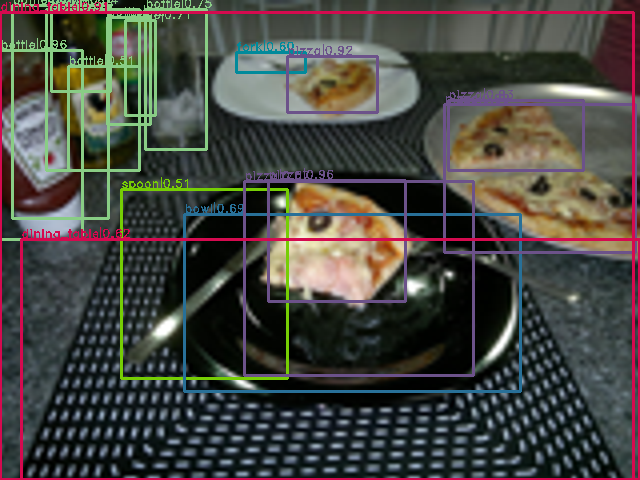}}\smallskip
		\end{minipage}
		\hfill
		\begin{minipage}[b]{0.24\linewidth}
			\centering
			\centerline{0.64/0.60}
			\centerline{\includegraphics[width=2.15cm]{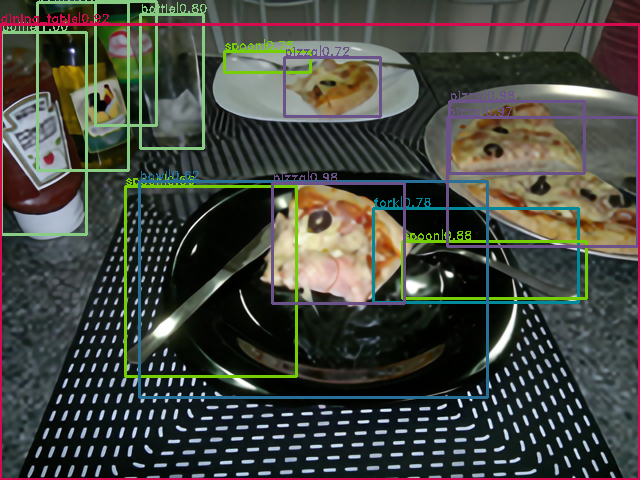}}\smallskip
		\end{minipage}
		\hfill
		\begin{minipage}[b]{0.24\linewidth}
			\centering
			\centerline{0.71/0.80}
			\centerline{\includegraphics[width=2.15cm]{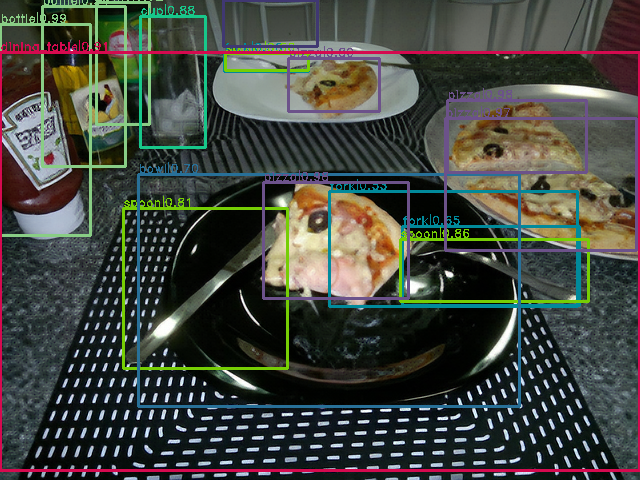}}\smallskip
		\end{minipage}
		\begin{minipage}[b]{0.24\linewidth}
			\centering
			\centerline{Precison/Recall}
			\centerline{\includegraphics[width=2.15cm]{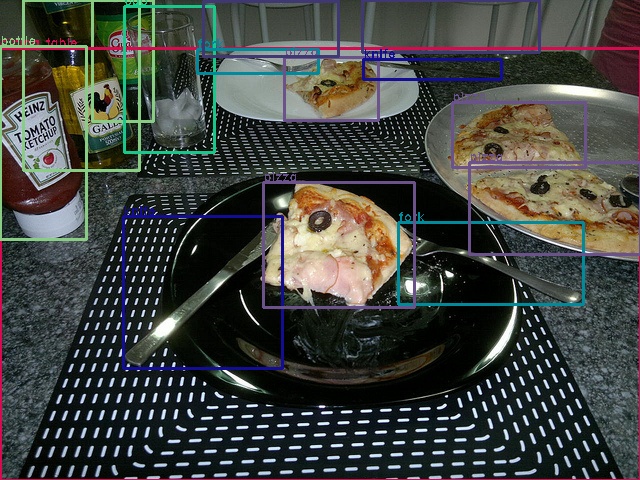}}\smallskip
		\end{minipage}
		\hfill

		\begin{minipage}[b]{0.24\linewidth}
			\centering
			\centerline{0.53/0.53}
			\centerline{\includegraphics[width=2.15cm]{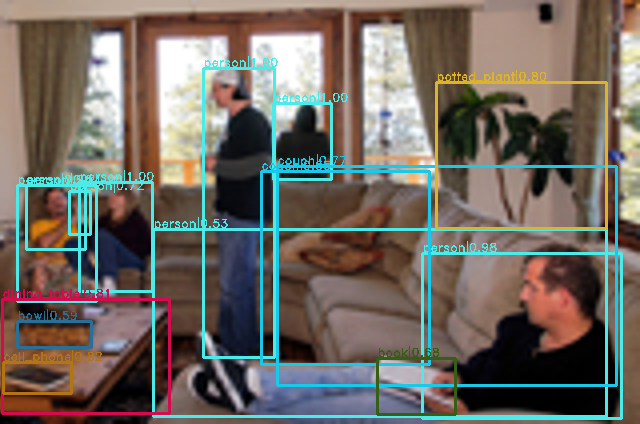}}\smallskip
			
		\end{minipage}
		\hfill
		\begin{minipage}[b]{0.24\linewidth}
			\centering
			\centerline{0.56/0.60}
			\centerline{\includegraphics[width=2.15cm]{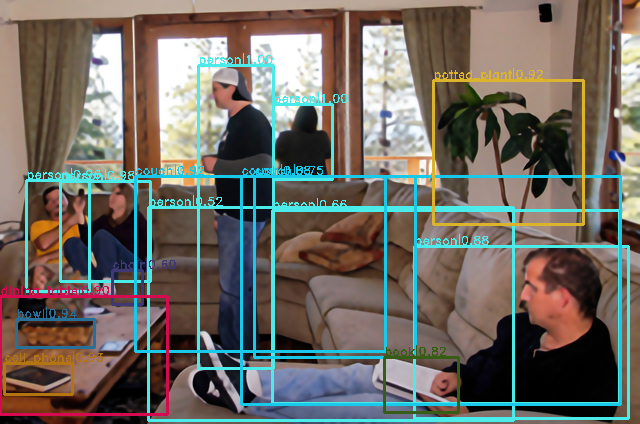}}\smallskip
		\end{minipage}
		\hfill
		\begin{minipage}[b]{0.24\linewidth}
			\centering
			\centerline{0.79/0.73}
			\centerline{\includegraphics[width=2.15cm]{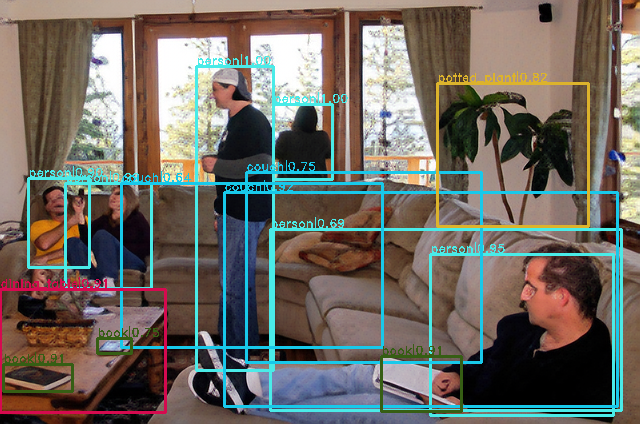}}\smallskip
		\end{minipage}
		\begin{minipage}[b]{0.24\linewidth}
			\centering
			\centerline{Precison/Recall}
			\centerline{\includegraphics[width=2.15cm]{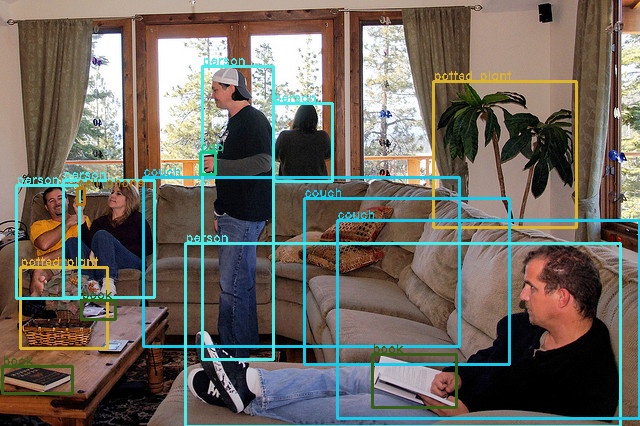}}\smallskip
		\end{minipage}
		\hfill

		\begin{minipage}[b]{0.24\linewidth}
			\centering
			\centerline{0.73/0.61}
			\centerline{\includegraphics[width=2.15cm]{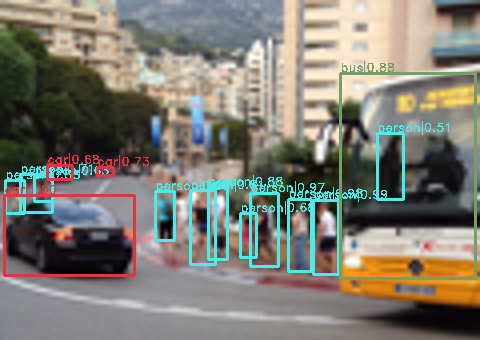}}\smallskip
			\centerline{(a) Bicubic}
		\end{minipage}
		\hfill
		\begin{minipage}[b]{0.24\linewidth}
			\centering
			\centerline{0.75/0.67}
			\centerline{\includegraphics[width=2.15cm]{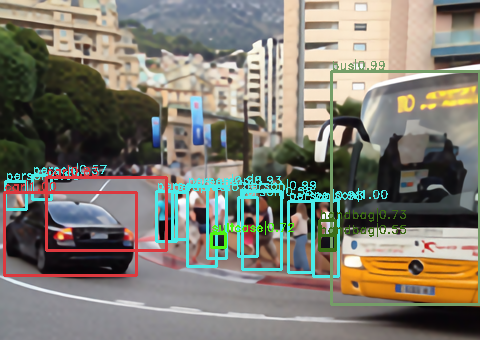}}\smallskip
			\centerline{(b) D-DBPN}
		\end{minipage}
		\hfill
		\begin{minipage}[b]{0.24\linewidth}
			\centering
			\centerline{0.87/0.72}
			\centerline{\includegraphics[width=2.15cm]{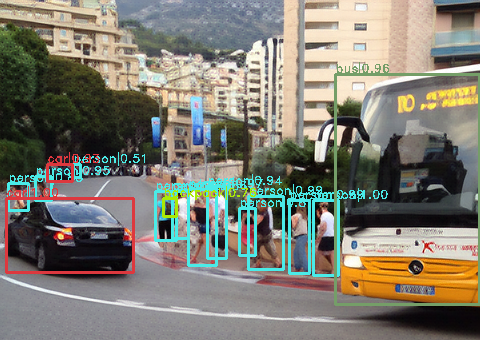}}\smallskip
			\centerline{(c) FDSR(ours)}
		\end{minipage}
		\begin{minipage}[b]{0.24\linewidth}
			\centering
			\centerline{Precison/Recall}
			\centerline{\includegraphics[width=2.15cm]{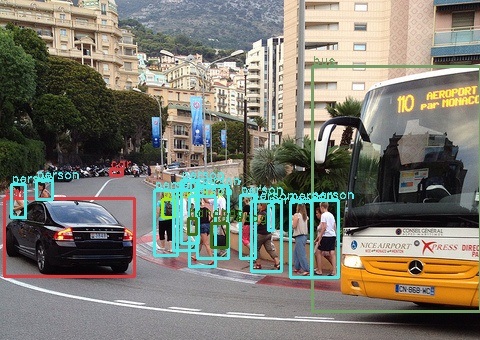}}\smallskip
			\centerline{(d) HR(GT)}
		\end{minipage}
		\hfill
		\vspace{-0.5cm}
		\caption{SR boosts object detection performance for low-resolution images, some examples of SR+detector (Faster R-CNN with FPN \cite{ mmdetection2019chen}) testing protocol on MS COCO 2017 validation dataset. (a) are the results  generated by bicubic $\times$4, (b) are the results generated by D-DBPN\cite{haris2018deep} $\times$4, (c) are the results from FDSR $\times$4. (d) are the ground-truth (GT) in HR image. Object detection index (precision and recall) are listed with the threshold of Intersection-over-Union (IoU) 0.5. }
		\label{fig:example}
		\vspace{-0.5cm}
	\end{figure}

Single image super-resolution (SISR) is an ill-posed inverse problem that tries to restore a high-resolution (HR) image from one or multiple low-resolution (LR) image(s).
The key of SISR relies on using efficient regularizer(s) as image priors~\cite{zhang2012single}.
Recently, there are plenty of works focused on giving simple, efficient and elegant solutions for solving this challenging problem~\cite{EDSR2017Lim, haris2018deep, zhang2018RCAN}.

Generally, current SISR methods can be divided into two categories from the view of ``consumer" of the algorithms: human-oriented and machine-oriented approaches.
Human-oriented methods aim at yielding visual pleasant images for people viewing and distinguishing. In order to render the missing high-frequent information due to image degradation, reconstruction indicators such as PSNR (peak signal-to-noise ratio) and SSIM (structure similarity) are widely used to guide the image reconstruction optimization process~\cite{EDSR2017Lim, haris2018deep, zhang2018RCAN}. Most of these methods do not consider the following specific tasks, for example, target detection, segmentation, and identification. Thus the results for there methods trend to pleasure human eyes.

\begin{figure*}[t]
	\vspace{-0.4cm}	
	\centering{\includegraphics[width=17.5cm,height=4.5cm]{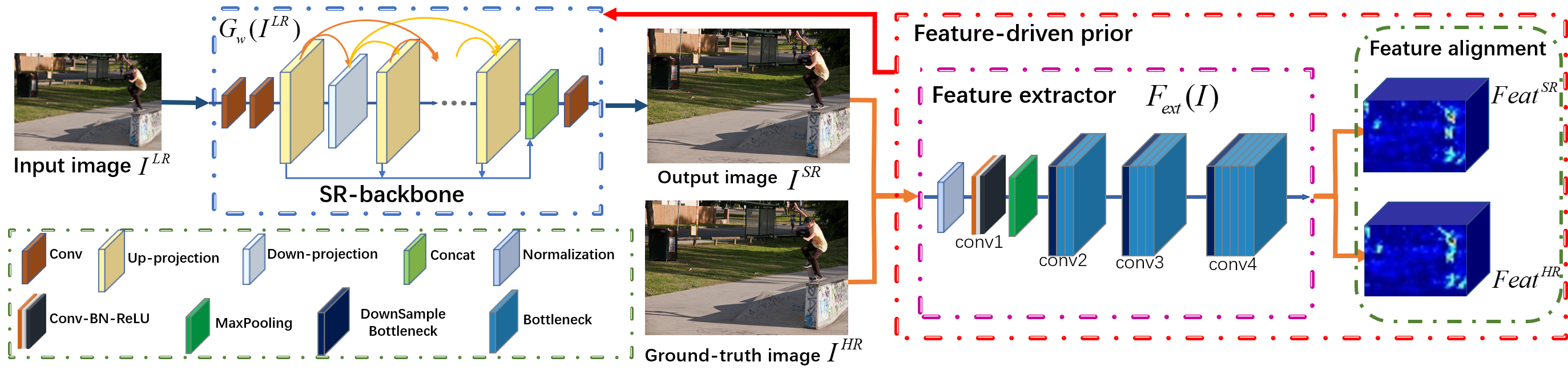}}
\vspace{-0.4cm}	
	\caption{The architecture of FDSR. Feature-drive prior is used to guide the reconstructing of SR by projecting super-resolved and ground-truth features from into a similar space for feature alignment. The feature extractor is the backbone of a well trained object detection model without retraining. }
	\label{fig:netarch}
\vspace{-0.5cm}
\end{figure*}

Machine-oriented methods consider the purpose of SISR's following task, which treats the SISR as a pre-processing. The design principles focus on learning resolution-invariant features for specific tasks to deal with multi-scale targets in one image.
Tan \emph{et al.}~\cite{tan2018feature} used Generative Adversarial Network (GAN) to super-resolved features for making machines see clearly. This method trended to generate enhanced features for image retrieval which depends on distinguishing features.
Haris \emph{et al.}~\cite{tdsr2018haris} claimed that a task-driven model has different effects on the SR network which using the high-level task to guide the network parameters optimization for adapting the cascaded specific tasks.
Dai \emph{et al.}~\cite{dai2016image} confirmed that SR is helpful for other vision tasks, even simply cascading SR without other tasks' specific needs (retraining), SR boosts the cascaded task performance.
Dong \emph{et al.}~\cite{dong2014learning} first introduced convolutional neural networks (CNNs) into SR problem.
Inspired by this pioneering work, various specific network architectures are designed to continuously improve the performance of SR \cite{EDSR2017Lim, haris2018deep, zhang2018RCAN}.
Furthermore, various vision-based tasks such as object detecting ~\cite{fasterrcnn2015ren, fpn2017lin, maskrcnn2017he, lin2017focal, cai2018cascade} used CNNs to train one model to solve a specific task in an end-to-end manner. Intuitively, twice training or multi-task learning schemes can boost both SR and cascaded task performance. However, retraining is time-consuming and how to design an effective network structure for multi-tasks is not a simple matter.
Thus, there is a problem worth noting how the SR really works for boosting other vision task performance.
Generally, vision-based tasks rely on clear and enough structure and texture information to extract specific features, SR gives more images details for clearly seeing by both humans and machines.

Based on a basic observation that both SR and detecting neural networks rely on learning specific features. We propose a novel feature-driven Super-Resolution (FDSR) scheme to provide a general solution for learning machine-oriented SR features.
First, FDSR relies on the joint guidance of feature-domain and pixel-domain loss to fine-tune the SR models.
Then, feature-driven prior is a plug and play manner that is compatible with other vision-based tasks without re-design the network.
Finally, we verify the effectiveness of FDSR in the object detection task in Fig.\ref{fig:example}.
The results show that the object detection results of using Faster R-CNN with FPN~\cite{fasterrcnn2015ren, fpn2017lin} on three SR algorithms of MS COCO 2017~\cite{lin2014microsoft} validation dataset. FDSR has better detection performance on Faster R-CNN with FPN, and the recovered image details are richer.
The main contributions of this work are:
(1) We propose a simple but powerful feature-driven SR scheme for reducing the feature gap between SR and cascaded tasks in a flexible and efficient manner.
(2) The plug and play nature of FDSR has generalizations to different detection networks.

	\section{Feature-driven super-resolution}
	\label{fdsr}
	
	FDSR relies on two parts: a super-resolution (SR) network and a feature extractor from one object detection network. The architecture of the network is shown in Fig.\ref{fig:netarch}. We use the backbone of the object detection model as a feature extractor to constrain the reconstructed features are as similar as possible with high-dimensional features from HR images. Generally, the two-step (SR+detector) model is easy to deploy without retaining and redesigning specific task networks. From this point, we propose a novel feature-driven SR method to provide enhance specific task performance which can extend the SR applications in real-world scenarios.
	To verify the validity of the proposed method, we use D-DBPN \cite{haris2018deep} as the SR backbone and Faster R-CNN with FPN \cite{fasterrcnn2015ren, fpn2017lin} as the cascaded detector for testing.
	In the test phase, we just use the two-step protocol. In the training phase, the feature extractor can be replaced with arbitrary existing detectors.
	Here, Faster R-CNN with FPN (one of the most powerful detectors) is used as the selected detector for testing. In the paper, detecting index is used as an evaluation indicator, from this point, FDSR has excellent compatibility with other detectors which means it has a strong generalization ability for different detectors.

	\vspace{-0.3cm}
	\subsection{Reconstruction Loss}
	
	Normally, Human-oriented SR methods are trained by using some kinds of reconstruction loss, such as L1 loss (i.e., mean absolute error) and L2 loss (i.e., mean square error) between HR and SR images. In this paper, we use D-DBPN as the backbone network to form the image reconstruction loss. D-DBPN\cite{haris2018deep} provides an up-to-down mapping unit to restore details through a reorganized feedback mechanism that alternates up- and down- sampling operations. Because D-DBPN achieves competitive results on standard SR benchmarks, we use D-DBPN to define reconstruction loss $L_p^{SR}$:
	
	\vspace{-0.3cm}
	\begin{equation}
	\label{l1loss}
	L_p^{Rec} = \frac{{\rm{1}}}{N}\sum\limits_{i = 1}^N {\mathop {\left\| {G_w{{\left( {{I_i^{LR}}} \right)}} - I_i^{HR}} \right\|}\nolimits_p } ,
	\vspace{-0.2cm}
	\end{equation}
where $I^{LR}$ is the LR image, $I^{HR}$ is the HR image and $G_w(I^{LR})$ is the image generated by the SR model (D-DBPN),  $w$ is the parameters of SR network, $i$ is the index of samples from training set, $N$ is the total number of samples, $p=1, 2$ represents different norms, here ($p = 1$).
	
	\vspace{-0.3cm}
	\subsection{Feature-driven Loss}
	
	Feature extraction is the most basic and important step in the target detection task. The quality of the extracted features directly affects the subsequent detection results.     
	In order to constrain the super-resolved image to shared similar features with the HR image, we derive a chain propagation rule for feature-driven training. Feature-driven manner only uses the features from detector without high-level semantic operations: prediction and localization.
	
	We use the backbone of Mask R-CNN with ResNet50-C4 \cite{maskrcnn2017he} as a feature extractor $F_{ext}$. On the one hand, Mask R-CNN introduces mask repercussion, the feature map extracted by its backbone is finer than Faster R-CNN. On the other hand, it reduces the excessive coupling between the feature extractor and subsequent detection networks, which is helpful to improve the usability of the generated image in other detection networks.
	Here, mean squared error (MSE) loss $L_{MSE}^{Feat}$ is used to describe the difference of feature extracted by the backbone between HR image and the reconstructed  image:    
	\vspace{-0.1cm}
	\begin{equation}\label{mseloss}
	L_{MSE}^{Feat} = \frac{1}{N}\sum\limits_{i = 1}^N {{{\left\| {F_{ext}{{\left( {G_w\left( {{I_i^{LR}}} \right)} \right)}} - F_{ext}{{\left( {{I_i^{HR}}} \right)}}} \right\|}_F}} ,
	\end{equation}
	where $F_{ext}\left( {G_w\left( {{I^{LR}}} \right)} \right)$ is the feature of reconstructed image, $F_{ext}\left( {{I^{HR}}} \right)$ is the feature of the ground-truth image, $F$ means Frobenius norm, $i$ is the index of images.
	
	\vspace{-0.3cm}
	\subsection{Optimization}
	The reconstruction loss can effectively maintain the structural information and texture information of the image, and the feature-domain loss can drive the SR network to learn better task-specific features which are the downstream network's main purpose.    
	We integrate these two losses as:
	
	\vspace{-0.2cm}
	\begin{equation}\label{comploss}
	L_{total} = \alpha L_p^{Rec} + \beta L_{MSE}^{Feat},
	\end{equation}
	where $\alpha$ and $\beta$ are weights determining the relative strength of the reconstruction loss and feature-driven loss.
	FDSR can be trained by optimizing the following objective function:
	
	\vspace{-0.6cm}
	\begin{equation}\label{optw}
	\begin{split}
	\arg \mathop {\min }\limits_w & \frac{{\rm{1}}}{N}\sum\limits_{i = 1}^N \alpha {{{\left\| {\mathop G\nolimits_w \left( {I_i^{LR}} \right) - I_i^{HR}} \right\|}_p}}  + \\	& \frac{1}{N}\sum\limits_{i = 1}^N \beta {{{\left\| {\mathop F\nolimits_{ext} \left( {G_w \left( {I_i^{LR}} \right)} \right) - \mathop F\nolimits_{ext} \left( {I_i^{HR}} \right)} \right\|}_F}} ,
	\end{split}
	\end{equation}
	explicitly, $F_{ext}()$ is a CNNs and it's frozen during training.
	Through back-propagation, the gradient information of $L_{MSE}^{Feat}$ will be passed layer by layer to $G_w \left( {I_i^{LR}} \right)$. Then, this pre-pixel gradient is combined with the pre-pixel gradient of $L_p^{Rec}$. The SR network's parameters $w$ are updated using standard back-propagation from the combined gradient, and the objective function is gradually decreased.


	\section{Experiments}
	\label{exper}
	
	\vspace{-0.2cm}
	\subsection{Implementation Details}
		
	\textbf{Datasets} We initialize all experiments with a D-DBPN model previously trained on the DIV2K dataset \cite{agustsson2017ntire}. The pre-trained model and training dataset are provided by the author of \cite{haris2018deep}.
	We use Faster R-CNN with FPN~\cite{fasterrcnn2015ren, fpn2017lin, mmdetection2019chen} pre-trained on the MS COCO2017 train dataset and publicly available as well.
	For easy to train, we randomly select 20k images as training samples whose length and width are both bigger than 384 pixels from MS COCO2017 train dataset.
	The MS COCO2017 validation dataset is selected as the testing set.
	The LR training and testing images are obtained by bicubic downsampling from the original HR image from the datasets with a particular scaling factor (i.e., 1/4 in our experiments, corresponding to 4$\times$ SR).
	
	\textbf{Training setting} All models are fine-tuned for $2\times 10^6$ iterations using learning rate as 1e-4 and mini-batch size as 8. We used Adam with $ \beta_1=0.9$, $\beta_2=0.999$ and $\epsilon=10^{-8}$ as optimizer. All experiments are performed on NVIDIA TITAN V GPUs with PyTorch.	

	\begin{figure}[h]
		\vspace{-0.4cm}
		\centering{\includegraphics[width=7.5cm,height=5cm]{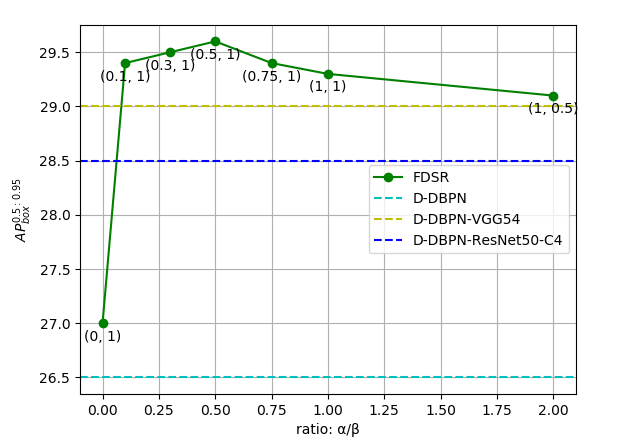}}
		\vspace{-0.4cm}
		\caption{The depth analysis of loss weight ($\alpha$, $\beta$) selection in FDSR. The effects of different types of feature-driven prior are also given. The $AP$ means the detection performance of Faster R-CNN on MS COCO validation dataset with 4$\times$ enlargement.}
		\label{fig:getbest}
		\vspace{-0.5cm}
	\end{figure}
	
	\vspace{-0.3cm}
	\subsection{The role of feature alignment}
	
	From Eq~\ref{comploss}, we can notice that the values of $\alpha$ and $\beta$ represent the contribution of different terms. Thus, we consider different settings of the values of $\alpha$ and $\beta$ to show the role of feature alignment.
	Due to the detecting oriented nature of the proposed method, we only use detection performance mean Average Precision (mAP) averaged for IoU $\in$ [0.5 : 0.05 : 0.95] (COCO’s standard metric, simply denoted $AP$) as the evaluation criteria. The downstream detector is selected as Faster R-CNN with FPN. The results are listed in Fig.\ref{fig:getbest}.

	
	The green line shows the performance of the SR images generated by FDSR with different settings of $\alpha$ and $\beta$, and the results show that $\alpha = 0.5$ and $\beta = 1$ is the better choice.
	The $AP$ = 26.5 of D-DBPN and the $AP$ = 27.0 of feature-driven only ($\alpha$ = 0, $\beta$ = 1) are both smaller than the $AP$ = 29.6 of FDSR, this phenomenon indicates that both of the reconstruction loss and the feature loss are important, however, for detecting performance consideration, feature-driven contribution is double than reconstruction contribution.
	Furthermore, considering the types of feature-driven prior, with the best performance weights configuration, we replace the feature-driven extractor as VGG19-54 (yellow dotted line) and ResNet50-C4 (blue dotted line) which are trained on ImageNet for the image classification task.
	FDSR outperforms some other kinds of feature-driven prior, this phenomenon testifies the important role of feature alignment with task relevance.
	
\begin{table}[]
	\caption{The PSNR(dB), FSIM of COCO 2017 VAL dataset generated by different state-of-art SR models. And detection performance (\%) of these images cascaded detector Faster R-CNN with FPN. The best score in each column is colored by red. AP$_S$, AP$_M$ and AP$_L$ represent small, middle and large scale target in images. }
	\centering
	\resizebox{\linewidth}{!}{
		\begin{tabular}{l|c|cccc|c}
			\hline
			Method & origin & Bicubic & EDSR \cite{EDSR2017Lim}   & RCAN \cite{zhang2018RCAN}  & D-DBPN \cite{haris2018deep} & FDSR \\ \hline
			PSNR   & \_     & 25.89                        & 28.17                     & {\color[HTML]{CB0000} 28.22} & 28.15  & 27.38                         \\
			FSIM   & \_     & 0.9872                       & 0.9936                    & 0.9936                       & 0.9938 & {\color[HTML]{CB0000} 0.9942} \\ \hline
			$AP$     & 37.7   & 22.3                         & 26.7                      & 26.5                         & 26.5   & {\color[HTML]{CB0000} 29.6}   \\
			AP$_{0.5}$   & 59.2   & 37.9                         & 44.5                      & 44.2                         & 44.3   & {\color[HTML]{CB0000} 49.1}   \\
			AP$_{0.75}$   & 41.1   & 23.0                         & 27.8                      & 27.5                         & 27.6   & {\color[HTML]{CB0000} 31.0}   \\ \hline
			AP$_S$    & 21.9   & 6.7                          & 8.8                       & 8.8                          & 9.0    & {\color[HTML]{CB0000} 11.2}   \\
			AP$_M$    & 41.4   & 23.8                         & 29.3                      & 29.0                         & 29.2   & {\color[HTML]{CB0000} 32.6}   \\
			AP$_L$    & 48.7   & 37.4                         & 42.3                      & 42.4                         & 42.5   & {\color[HTML]{CB0000} 44.6}   \\ \hline
	\end{tabular}}
	\label{srresult}
	\vspace{-0.4cm}
\end{table}

\vspace{-0.3cm}	
\subsection{Comparison with state-of-the-art SR methods}
	
Some state-of-the-art SR methods: EDSR \cite{EDSR2017Lim}, D-DBPN \cite{haris2018deep}, RCAN \cite{zhang2018RCAN} with 4$\times$ scale factor are selected as the benchmarks. The detection performance of these generated images are tested on MS COCO 2017 validation dataset
Table.~\ref{srresult} shows the results regarding as PSNR, FSIM, and detection performance.
From the experimental results, we find that the SR has a drastic effect on the detection results of Faster R-CNN. For $4\times$ downsampling, the $AP$ dropped from 37.7\% to 22.3\%.
Among them, small (area $< 32^2$) and medium ($32^2 \le$ area $< 96^2$) targets are more severely affected, AP$_S$ drop from 21.9\% to 6.7\%, AP$_M$ drop from 41.4\% to 23.8\%.
It is speculated that this is due to the actual loss of information and the limitations of the detector architecture. The performance of SR cannot be significantly improved by using the pixel domain loss-driven SR method, the weak correlation between PSNR and detection performance also supports this view.
Apart from PSNR, there is a stronger correlation between FSIM and detection performance, which also shows the importance of enhanced features from high-level downstream tasks. The proposed FDSR achieves significantly better results, compared with the official D-DBPN, the $AP$ increase by 3.1 points on the detection results.

\begin{table}[]
	\caption{FDSR detection performance on COCO 2017 VAL dataset on 4$\times$ scale factor cascaded with different detectors.}
	\centering
	\resizebox{\linewidth}{!}{
		\begin{tabular}{l|l|cccc}
			\hline
			\multirow{2}{*}{Methods} & \multirow{2}{*}{Backbone} & \multicolumn{4}{c}{$AP$}                                             \\ \cline{3-6}
			&                           & {HR} & {Bicubic} & {D-DBPN} & FDSR \\ \hline
			Faster-RCNN            & R101-FPN            & 39.4                    & 25.2                         & 28.2                        & 31.4 \\
			RetinaNet               & R50-FPN             & 36.4                    & 22.2                         & 25.7                        & 28.6 \\
			Cascade-RCNN            & R50-FPN             & 40.4                    & 24.5                         & 28.4                        & 31.7 \\ \hline
	\end{tabular}}
	\label{other-result}
	\vspace{-0.4cm}
\end{table}

\vspace{-0.3cm}
\subsection{Adaptability of FDSR for other detectors}
\vspace{-0.1cm}
	In order to verify the wide applicability of FDSR in detection tasks, we conduct testing with different detectors using different backbones (Faster RCNN-Resnet101-FPN~\cite{fasterrcnn2015ren}, RetinaNet-Resnet50-FPN~\cite{lin2017focal}, Cascade R-CNN-ResNet50-FPN~\cite{cai2018cascade}). The experimental results are shown in Table.~\ref{other-result}. Regardless of which detectors, comparing with benchmark SR method D-DBPN, the detecting performances are significantly improved by FDSR. This further confirms the good adaptability of FDSR combined with other detectors.

	\vspace{-0.3cm}
	\subsection{Difference with TDSR}
	\vspace{-0.1cm}
	
	\begin{table}[]
		\caption{ The PSNR(dB), FSIM of VOC2007 testing dataset with 4$\times$ scale factor by different SR methods. In line with TDSR, the detector is selected as SSD~\cite{zhang2012single}. Original HR images obtain 77.43\% mAP.}
		\label{result-voc}
		\centering
		\begin{tabular}{l|c|c|c}
			\hline
			Method  & PSNR  & FSIM   & mAP   \\ \hline
			Bicubic & 25.95 & 0.9656 & 48.85 \\
			D-DBPN \cite{haris2018deep}  & 28.42 & 0.9858 & 59.81 \\
			TDSR \cite{tdsr2018haris}    & 27.49 & 0.9889 & 67.96  \\ \hline
			FDSR (ours)    & 27.51 & 0.9889 & 67.98 \\ \hline
		\end{tabular}
\vspace{-0.4cm}
	\end{table}
In order to show the difference between FDSR and TDSR, we also evaluate all selected SR methods on PASCAL VOC 2007 test dataset~\cite{everingham2010voc} with TDSR 4$\times$~\cite{tdsr2018haris}.
TDSR optimizes D-DBPN to improve the performance of object detection by using the task loss on the SSD network. The task loss includes the classification loss and bounding box regression loss between detection results and ground-truth, which guides the SR model to optimize for a specific detection network. In addition, the structure of TDSR is so deep that it may cause the problem of gradient disappearance. For fairness, the dataset processing method is consistent with TDSR that LR images are obtained by bicubic downscaling the original (HR, 300$\times$300 pixels) image from the data set with 4$\times$ factor. We reproduce TDSR and get the experimental results. The results are provided in Table.~\ref{result-voc}. Compared with TDSR, our method has slightly better performance on detection performance without fine-tune on the VOC dataset. This also shows that FDSR has strong universality on different datasets and object detection networks.

	\vspace{-0.3cm}
	\section{Conclusions}
	\label{conclu}
	\vspace{-0.1cm}
	Aiming at reducing the task gap between SR network and other high-level tasks, we summarize the advantages and disadvantages of the existing SR methods to propose a flexible and efficient framework named Feature-Driven Super-Resolution.
FDSR directly use task-oriented feature-drive prior to maintain the feature consistency between the generated super-resolved images and HR image. We verify the effectiveness of the proposed method in the target detection task, FDSR has achieved significant improvements in different detectors.
	
	\bibliographystyle{IEEEbib}
	\bibliography{refs} 
	
\end{document}